\documentclass[journal]{IEEEtran}

\usepackage{amsmath,amssymb}
\usepackage{algorithm,algpseudocode}
\usepackage{booktabs}
\usepackage{color}
\usepackage{courier}
\usepackage{epsfig}
\usepackage{graphicx}
\usepackage{helvet}
\usepackage{multirow}
\usepackage{times}
\usepackage{url}
\usepackage{cite}

\newcommand\etal{{et~al.}}
\newcommand\ie{{i.e.}}
\newcommand\etc{{etc.}}
\newcommand\eg{{e.g.}}

\ifCLASSINFOpdf
\else
\fi

\begin{document}

\title{Arbitrary-Oriented Scene Text Detection via Rotation Proposals}

\author{Jianqi Ma, Weiyuan Shao, Hao Ye, Li Wang, Hong Wang, Yingbin Zheng, Xiangyang Xue
\thanks{J. Ma, L. Wang, and X. Xue are with Shanghai Key Lab of Intelligent Information Processing, School of Computer Science, Fudan University, Shanghai 200433, China. W. Shao, H. Ye, H. Wang, and Y. Zheng are with Shanghai Advanced Research Institute, Chinese Academy of Sciences, Shanghai 201210, China. The first two authors contributed equally to this work. Corresponding author: Yingbin Zheng (e-mail: zhengyb@sari.ac.cn).}
}

\maketitle

\begin{abstract}
This paper introduces a novel rotation-based framework for arbitrary-oriented text detection in natural scene images. We present the \emph{Rotation Region Proposal Networks} (RRPN), which are designed to generate inclined proposals with text orientation angle information. The angle information is then adapted for bounding box regression to make the proposals more accurately fit into the text region in terms of the orientation. The \emph{Rotation Region-of-Interest} (RRoI) pooling layer is proposed to project arbitrary-oriented proposals to a feature map for a text region classifier. The whole framework is built upon a region-proposal-based architecture, which ensures the computational efficiency of the arbitrary-oriented text detection compared with previous text detection systems. We conduct experiments using the rotation-based framework on three real-world scene text detection datasets and demonstrate its superiority in terms of effectiveness and efficiency over previous approaches.
\end{abstract}

\begin{IEEEkeywords}
Scene text detection, arbitrary oriented, rotation proposals.
\end{IEEEkeywords}

\IEEEpeerreviewmaketitle

\section{Introduction}
\label{sec:intro}

Text detection aims to identify text regions of given images and is an important prerequisite for many multimedia tasks, such as visual classification \cite{2017tmm_SKaraoglu,2017arxiv_xbai}, video analysis \cite{2016tip_xyin,2012tmm_xliu} and mobile applications \cite{2011tmm_kbouman}. Although there are a few commercial optical character recognition (OCR) systems for documentary texts or internet content, the detection of text in a natural scene image is challenging due to complex situations such as uneven lighting, blurring, perspective distortion, orientation, \etc

In recent years, much attention has been paid to the text detection task (\eg,
\cite{CVPR2004Chen,wang2010word,ACCV2011Neumann,ICCV2013PhotoOCR,Jaderberg2014Deep,Huang2014Robust,2015tmm_lwu,Tian2015Text,BazazianGNBKB17,2017tmm_xren,Liao2016TextBoxes}). Although these approaches have shown promising results, most of them rely on horizontal or nearly horizontal annotations and return the detection of horizontal regions. However, in real-world applications, a larger number of the text regions are not horizontal, and even applying non-horizontal aligned text lines as the axis-aligned proposals may not be accurate. Thus, the horizontal-specific methods cannot be widely applied in practice.

Recently, a few works have been proposed to address arbitrary-oriented text detection \cite{zhang2016multi,yao2016scene,he2016accurate}. In general, these methods mainly involve two steps, \ie, segmentation networks, such as the fully convolutional network (FCN), are used to generate text prediction maps, and geometric approaches are used for inclined proposals. However, prerequisite segmentation is usually time-consuming. In addition, some systems require several post-processing steps to generate the final text region proposals with the desired orientation and are thus not as efficient as those directly based on a detection network.

\begin{figure}[t]
  \centering
  \includegraphics[width=\linewidth]{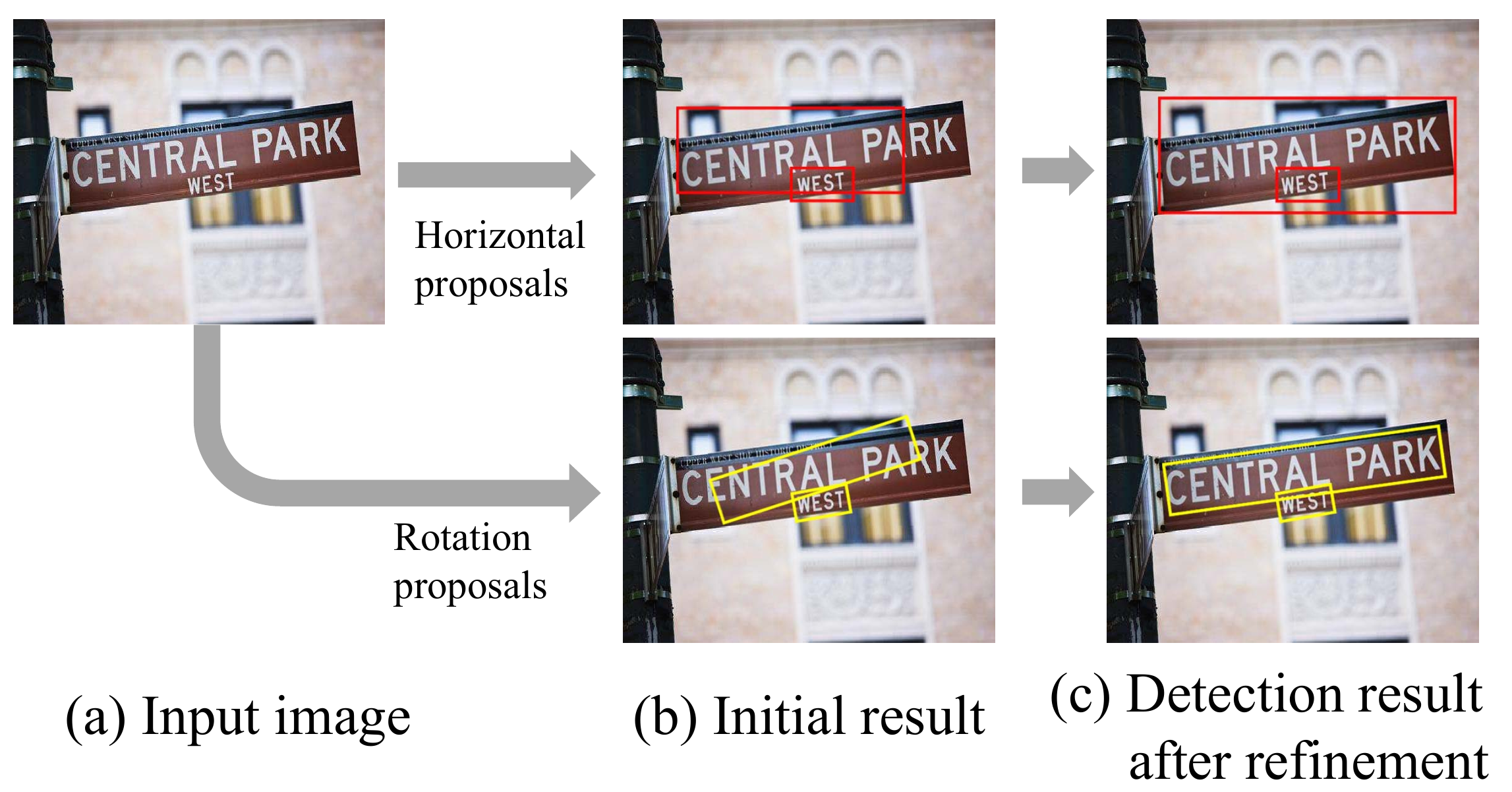}
  \caption{Overview of text detection. First row: text detection based on horizontal bounding box proposal and bounding box regression of Faster-RCNN \cite{ren2016Faster}. Second row: detection using rotation region proposal and bounding box regression with orientation step.}
  \label{fig:overview}
\end{figure}

In this paper, we develop a rotation-based approach and an end-to-end text detection system for arbitrary-oriented text detection. Particularly, orientations are incorporated so that the detection system can generate proposals for arbitrary orientation. A comparison between the previous horizontal-based approach and ours is illustrated in Figure \ref{fig:overview}. We present the \emph{Rotation Region Proposal Networks} (RRPN), which are designed to generate inclined proposals with text orientation angle information. The angle information is then adapted for bounding box regression to make the proposals more accurately fit the text region. The \emph{Rotation Region-of-Interest} (RRoI) pooling layer is proposed to project arbitrary-oriented proposals to a feature map. Finally, a two-layer network is deployed to classify the regions as either text or background. The main contributions of this paper include the following:
\begin{itemize}
\item Different from previous segmentation-based frameworks, ours has the ability to predict the orientation of a text line using a region-proposal-based approach; thus, the proposals can better fit the text region, and the ranged text region can be easily rectified and is more convenient for text reading. New components, such as the RRoI pooling layer and learning of the rotated proposal, are incorporated into the region-proposal-based architecture \cite{ren2016Faster}, which ensures the computational efficiency of text detection compared with segmentation-based text detection systems.
\item We also propose novel strategies for the refinement of region proposals with arbitrary orientation to improve the performance of arbitrary-oriented text detection.
\item We apply our framework to three real-world text detection datasets, \ie, MSRA-TD500 \cite{Yao2012Detecting}, ICDAR2013 \cite{Karatzas2013ICDAR} and ICDAR2015 \cite{karatzas2015icdar}, and find that it is more accurate and significantly efficient compared to previous approaches.
\end{itemize}

The rest of this paper is organized as follows. Section \ref{sec:related} introduces the background of scene text detection and related work. Section \ref{sec:hrp} briefly reviews the horizontal region proposal approach. Section \ref{sec:app} discusses our framework in detail. In Section \ref{sec:evaluation}, we demonstrate the quantitative study on three datasets. We conclude our work in Section \ref{sec:conclusion}.

\section{Related Work}
\label{sec:related}

The reading of text in the wild has been studied over the last few decades; comprehensive surveys can be found in \cite{chen2000survey,Jung2004Text,Uchida2014Text,Ye2015Text}.
Methods based on the sliding window, connected components and the bottom-up strategy are designed to handle horizontal-based text detection.
Sliding window-based methods \cite{kim2003texture,chen2004detecting,wang2010word,neumann2013scene,Jaderberg2014Deep} tend to use a sliding window of a fixed size to slide the text area and find the region most likely to include text. To consider more precise styles of text, \cite{Jaderberg2014Deep,Wang2012End} apply multiple scales and ratios to the sliding window methods. However, the sliding window process leads to a large computational cost and inefficiency. Representative connected-component-based approaches such as the Stroke Width Transform (SWT) \cite{CVPR2010SVT} and Maximally Stable Extremal Regions (MSER) \cite{Matas2004Robust} exhibited superior performances in the ICDAR 2011 \cite{Karatzas2011ICDAR} and ICDAR 2013 \cite{Karatzas2013ICDAR} robust text detection competitions. They mainly focus on the edge and pixel point of an image by detecting the character via edge detection or extreme region extraction and then combining the sub-MSER components into a word or text-line region. The capabilities of these methods are limited in some difficult situations involving multiple connected characters, segmented stroke characters and non-uniform illumination \cite{zhang2016character}.

Scene text in the wild is usually aligned from any orientation in real-world applications, and approaches for arbitrary orientations are needed. For example, \cite{Risnumawan2014A} uses mutual magnitude symmetry and gradient vector symmetry to identify text pixel candidates regardless of the orientation, including curves from natural scene images, and \cite{Cho_2016_CVPR} designs a Canny text detector by taking the similarity between an image edge and text to detect text edge pixels and perform text localization. Recently, convolution-network-based approaches were proposed to perform text detection, \eg, Text-CNN \cite{he2016text}, by first using an optimized MSER detector to find the approximate region of the text and then sending region features into a character-based horizontal text CNN classifier to further recognize the character region. In addition, the orientation factor is adopted in the segmentation models developed by Yao \etal \cite{yao2016scene}. Their model aims to predict more accurate orientations via an explicit manner of text segmentation and yields outstanding results on the ICDAR2013 \cite{Karatzas2013ICDAR}, ICDAR2015 \cite{karatzas2015icdar} and MSRA-TD500 \cite{Yao2012Detecting} benchmarks.

A technique similar to text detection is generic object detection. The detection process can be made faster if the number of proposals is largely reduced. There is a wide variety of region proposal methods, such as Edge Boxes \cite{Zitnick2014Edge}, Selective Search \cite{Uijlings2013Selective}, and Region Proposal Networks (RPNs) \cite{ren2016Faster}. For example, Jaderberg \etal \cite{Jaderberg2014Reading} extends the region proposal method and applies the Edge Boxes method \cite{Zitnick2014Edge} to perform text detection. Their text spotting system achieves outstanding results on several text detection benchmarks. The Connectionist Text Proposal Network (CTPN) \cite{tian2016detecting} is also a detection-based framework for scene text detection. It employs the image feature from the CNN network in LSTM to predict the text region and generate robust proposals.

This work is inspired by the RPN detection pipeline in regards to the dense-proposal based approach used for detection and RoI pooling operation used to further accelerate the detection pipeline. Detection pipelines based on RPN are widely used in various computer vision applications \cite{Zhong2016DeepText,DBLP:journals/corr/JiangL16a,wang2017evolving}. The idea is also similar to that of Spatial Transformer Networks (STN) \cite{Jaderberg2015Spatial}, \ie, a neural network model can rectify an image by learning its affine transformation matrix. Here, we try to extend the model to multi-oriented text detection by injecting angle information. Perhaps the work most related to ours is \cite{Zhong2016DeepText}, where the authors proposed an inception-RPN and made further text detection-specific optimizations to adapt the text detection. We incorporate the rotation factor into the region proposal network so that it is able to generate arbitrary-oriented proposals. We also extend the RoI pooling layer into the Rotation RoI (RRoI) pooling layer and apply angle regression in our framework to perform the rectification process and finally achieve outstanding results.

\begin{figure*}[t]
  \centering
  \includegraphics[width=0.8\linewidth]{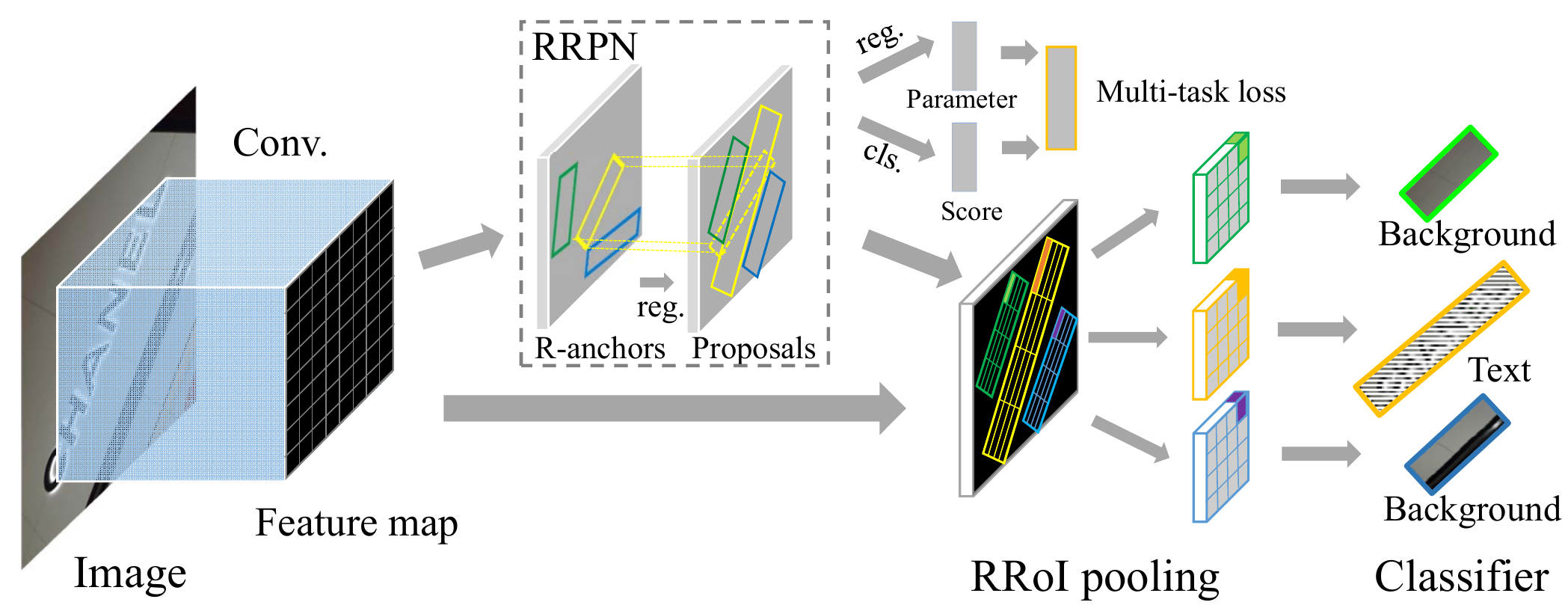}
  \caption{Rotation-based text detection pipeline.}
  \label{fig:framework}
\end{figure*}

\section{Horizontal Region Proposal}
\label{sec:hrp}

We begin with a brief review of RPN \cite{ren2016Faster}. As mentioned in the previous section, an RPN is able to further accelerate the process of proposal generation. Part of VGG-16 \cite{Simonyan2015Very} is employed as sharable layers, and the horizontal region proposals are generated by sliding over the feature map of the last convolutional layer. The features extracted from each sliding window are fed into two sibling layers (a box-regression (\emph{reg}) layer and a box-classification (\emph{cls}) layer), with 4$k$ (4 coordinates per proposal) outputs from the \emph{reg} layer representing coordinates and 2$k$ (2 scores per proposal) scores from the \emph{cls} layer for $k$ anchors of each sliding position.

To fit the objects to different sizes, the RPN uses two parameters to control the size and shape of anchors, \ie, scale and aspect ratio. The scale parameter determines the size of the anchor, and the aspect ratio controls the ratio of the width to the height for the anchor box. In \cite{ren2016Faster}, the authors set the scale as 8, 16 and 32 and the ratio as 1:1, 1:2 and 2:1 for a generic object detection task. This anchor selection strategy can cover the shapes of nearly all natural objects and keep the total number of proposals low. However, in the text detection task, especially for scene images, texts are usually presented in an unnatural shape with different orientations; axis-aligned proposals generated by RPN are not robust for scene text detection. To make a network more robust for text detection and maintain its efficiency, we think that it is necessary to build a detection framework, which encodes the rotation information with the region proposals.

\section{Approach}
\label{sec:app}

We now elaborate the construction of the rotation-based framework; the architecture is illustrated in Figure \ref{fig:framework}. We employ the convolutional layers of VGG-16 \cite{Simonyan2015Very} in the front of the framework, which are shared by two sibling branches, \ie, the RRPN and a clone of the feature map of the last convolutional layer. The RRPN generates arbitrary-oriented proposals for text instances and further performs bounding box regression for proposals to better fit the text instances. The sibling layers branching out from the RRPN are the classification layer (\emph{cls}) and the regression layer (\emph{reg}) of the RRPN. The outputs from these two layers are the scores from the \emph{cls} and proposal information from the \emph{reg}, and their losses are computed and summed to form a multitask loss. Then, the RRoI pooling layer acts as a max pooling layer by projecting arbitrary-oriented text proposals from the RRPN onto the feature map. Finally, a classifier formed by two fully connected layers is used, and the region with the RRoI features is classified as either text or background.

\subsection{Rotated Bounding Box Representation}

In the training stage, the ground truth of a text region is represented as rotated bounding boxes with 5 tuples $(x, y, h, w, \theta)$. The coordinate $(x, y)$ represents the geometric center of the bounding box. The height $h$ is set as the short side of the bounding box, and the width $w$, as the long side. The orientation $\theta$ is the angle from the positive direction of the x-axis to the direction parallel to the long side of the rotated bounding box. Because of the special ability of scene text detection, the direction of reading and its opposite do not influence the detected region. Here, we simply maintain the orientation parameter $\theta$ such that it covers half the angular space. Suppose the orientation of a rotated box is $\theta$; there exists one and only one integer $k$ ensuring that $\theta+k\pi$ is within the interval $[-\frac{\pi}{4}, \frac{3\pi}{4})$, and we update $\theta+k\pi$ as $\theta$. There are three benefits of the tuple representation $(x,y,h,w,\theta)$. First, it is easy to calculate the angle difference between two different rotated boxes. Second, this is a rotation-friendly representation for the angle regression of each rotated bounding box. Third, compared with the traditional 8-point representation ($x_1,y_1,x_2,y_2,x_3,y_3,x_4,y_4$) of a rotated bounding box, this representation can be used to easily calculate the new ground truth after we rotate a training image.

Suppose the size of a given image is $I_H\times I_W$ and the original text region is represented as $(x,y,h,w,\theta)$. If we rotate the image by an angle $\alpha\in [0,2\pi)$ around its center, the center of the anchor can be calculated as
\begin{equation}
    \left[
\begin{array}{c}
x'\\y'\\1
\end{array} \right]
=\mathbf{T}(\frac{I_W}{2},\frac{I_H}{2})\mathbf{R}(\alpha)\mathbf{T}(-\frac{I_W}{2},-\frac{I_H}{2})
    \left[
\begin{array}{c}
x\\y\\1
\end{array} \right]
\end{equation}
where $\mathbf{T}$ and $\mathbf{R}$ are the translation matrix and rotation matrix, respectively,
\begin{equation}
    \mathbf{T}(\delta_x,\delta_y)=    \left[
\begin{array}{lll}
1 & 0 & \delta_x\\
0 & 1 & \delta_y\\
0 & 0 & 1
\end{array} \right]
\end{equation}
\begin{equation}
    \mathbf{R}(\alpha)=    \left[
\begin{array}{ccc}
\cos\alpha & \sin\alpha & 0\\
-\sin\alpha & \cos\alpha & 0\\
0 & 0 & 1
\end{array} \right]
\end{equation}
The width $w'$ and height $h'$ of the rotated bounding box do not change, and the orientation is $\theta'=\theta + \alpha + k\pi$ ($\theta'\in [-\frac{\pi}{4}, \frac{3\pi}{4})$). We employed this image rotation strategy for data augmentation during training.

\subsection{Rotation Anchors}

\begin{figure}[t]
  \centering
  \includegraphics[width=.77\linewidth]{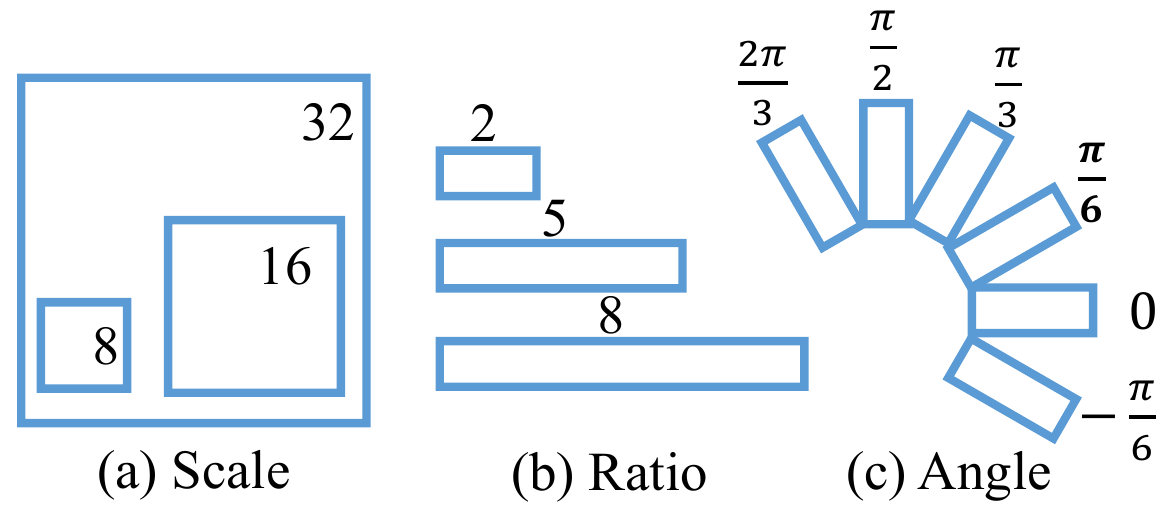}
  \caption{Anchor strategy used in our framework.}
  \label{fig:anchor}
\end{figure}

Traditional anchors, which use scale and aspect ratio parameters, are not sufficient for in-the-wild text detection. Therefore, we design the rotation anchors (R-anchors) by making several adjustments. First, an orientation parameter is added to control the orientation of a proposal. Six different orientations, \ie, $-\frac{\pi}{6}$, 0, $\frac{\pi}{6}$, $\frac{\pi}{3}$, $\frac{\pi}{2}$ and $\frac{2\pi}{3}$, are used, which are trade-offs between orientation coverage and computational efficiency. Second, as text regions usually have special shapes, the aspect ratio is changed to 1:2, 1:5 and 1:8 to cover a wide range of text lines. In addition, the scales of 8, 16 and 32 are kept. The anchor strategy is summarized in Figure \ref{fig:anchor}. Following our data representation step, a proposal is generated from the R-anchors with 5 variables ($x$, $y$, $h$, $w$, $\theta$). For each point on the feature map, 54 R-anchors (6 orientations, 3 aspect ratios, and 3 scales) are generated, as well as 270 outputs (5$\times$54) for the \emph{reg} layer and 108 score outputs (2$\times$54) for the \emph{cls} layer at each sliding position. Then, we slide the feature map with the RRPN and generate $H \times W \times 54$ anchors in total for the feature map, with width $W$ and height $H$.

\subsection{Learning of Rotated Proposal}

\begin{figure}[t]
  \centering
  \includegraphics[width=\linewidth]{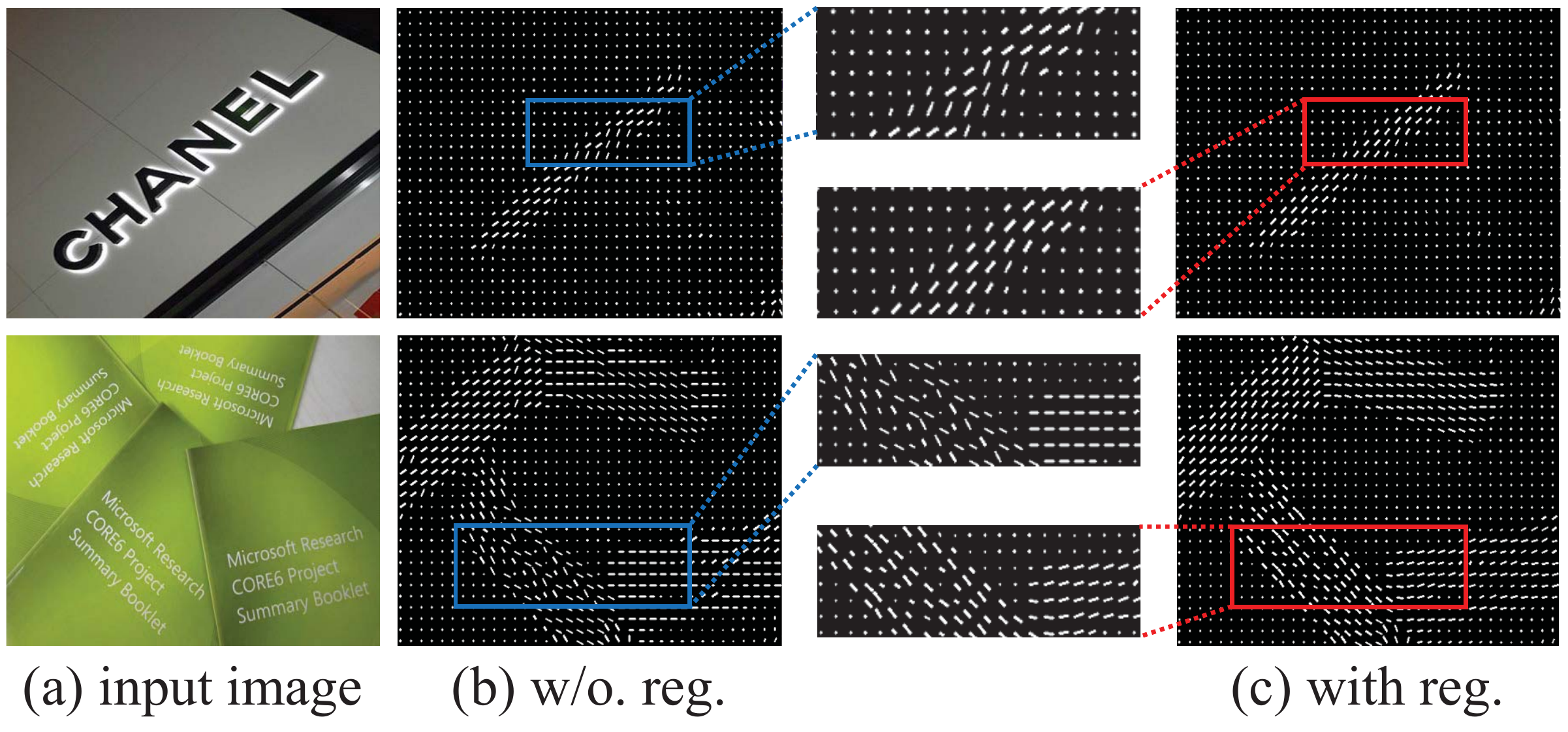}
  \caption{Visualization of the impact on regression: input images (a); orientation and response of the anchors without regression term (b) and with regression (c). The orientation of the R-anchor is the direction of the white line at each point, with longer lines indicating a higher response score for text.}
  \label{fig:RegVis}
\end{figure}

\begin{figure}[t]
  \centering
  \includegraphics[width=\linewidth]{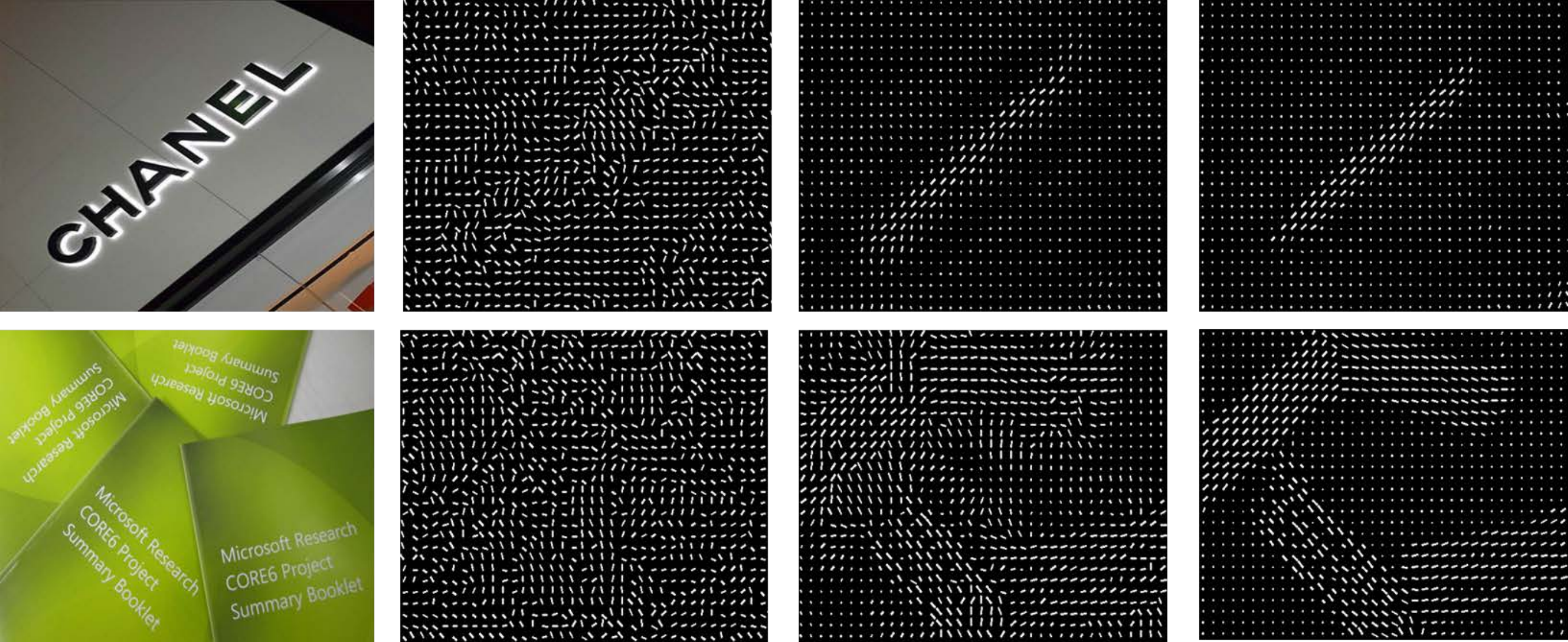}
  (a)~~~~~~~~~~~~~~(b)~~~~~~~~~~~~~~(c)~~~~~~~~~~~~~~(d)
  \caption{Visualization of different multitask loss values. (a) Input images; (b) 0 iterations; (c) 15,000 iterations; (d) 150,000 iterations.}
  \label{fig:loss}
\end{figure}

As the R-anchors are generated, a sampling strategy for the R-anchors is needed to perform network learning. We first define the intersection-over-union (IoU) overlap as the overlap between the skew rectangles of the ground truth and R-anchor. Then, positive R-anchors feature the following: (i) the highest IoU overlap or an IoU larger than 0.7 with respect to the ground truth, and (ii) an intersection angle with respect to the ground truth of less than
 $\frac{\pi}{12}$. Negative R-anchors are characterized by the following: (i) an IoU lower than 0.3, or (ii) an IoU larger than 0.7 but with an intersection angle with a ground truth larger than $\frac{\pi}{12}$. Regions that are not selected as either positive or negative are not used during training.

Our loss function for the proposal takes the form of multitask loss \cite{girshick2015fast}, which is defined as:
\begin{equation}
    L(p, l, v^{*}, v) = L_{\text{cls}}(p, l) + \lambda l L_{\text{reg}}(v^{*}, v)
\end{equation}
where $l$ is the indicator of the class label ($l=1$ for text and $l=0$ for background; no regression for the background), the parameter $p = (p_0, p_1)$ is the probability over classes computed by the softmax function, $v = (v_x, v_y, v_h, v_w, v_\theta)$ denotes the predicted tuple for the text label, and $v^{*} = (v^{*}_x, v^{*}_y, v^{*}_h, v^{*}_w, v^{*}_\theta)$ denotes the ground truth. The trade-off between two terms is controlled by the balancing parameter $\lambda$. We define the classification loss for class $l$ as:
\begin{equation}
    L_{\text{cls}}(p, l) = -\log{p_l}
\end{equation}
For the bounding box regression, the background RoIs are ignored, and we adopt smooth-$L_1$ loss for the text RoIs:
\begin{equation}
    L_{\text{reg}}(v^{*}, v) = \sum_{i\in\{x,y,h,w,\theta\}}{\text{smooth}_{L_1}(v_i^{*} - v_i)}
\end{equation}
\begin{equation}
\text{smooth}_{L_1}(x)=\left\{
\begin{array}{ll}
0.5x^2 & \text{if~} |x| < 1\\
|x| - 0.5 & \text{otherwise}
\end{array} \right.
\end{equation}

The scale-invariant parameterizations tuple $v$ and $v^*$ are calculated as follows:
\begin{equation}
\begin{array}{l}
    v_{x}=\frac{x - x_a}{w_a}, v_{y} = \frac{y-y_a}{h_a} \\
    v_{h}=\log\frac{h}{h_a},v_{w} = \log\frac{w}{w_a}, v_{\theta}=\theta \ominus \theta_a
\end{array}
\end{equation}
\begin{equation}
\begin{array}{l}
    v^*_{x}=\frac{x^* - x_a}{w_a}, v^*_{y} = \frac{y^*-y_a}{h_a} \\
    v^*_{h}=\log\frac{h^*}{h_a},v^*_{w} = \log\frac{w^*}{w_a}, v^*_{\theta}=\theta^* \ominus \theta_a
\end{array}
\end{equation}
where $x$, $x_a$ and $x^*$ are for the predicted box, anchor and ground truth box, respectively; the same is for $y$, $h$, $w$ and $\theta$. The operation $a \ominus b = a - b + k\pi$, where $k \in Z$ to ensure that $a \ominus b\in[-\frac{\pi}{4},\frac{3\pi}{4})$.

As described in the previous section, we give R-anchors fixed orientations within the range $[-\frac{\pi}{4}, \frac{3\pi}{4})$, and each of the 6 orientations can fit the ground truth that has an intersection angle of less than $\frac{\pi}{12}$. Thus, every R-anchor has its fitting range, which we call its fit domain. When an orientation of a ground truth box is in the fit domain of an R-anchor, this R-anchor is most likely to be a positive sample of the ground truth box. As a result, the fit domains of the 6 orientations divide the angle range $[-\frac{\pi}{4}, \frac{3\pi}{4})$ into 6 equal parts. Thus, a ground truth in any orientation can be fitted with an R-anchor of the appropriate fit domain. Figure \ref{fig:RegVis} shows a comparison of the utility of the regression terms. We can observe that the orientations of the regions are similar in a neighborhood region.

To verify the ability of a network to learn the text region orientation, we visualize the intermediate results in Figure \ref{fig:loss}. For an input image, the feature maps of RRPN training after different iterations are visualized. The short white line on the feature map represents the R-anchor with the highest response to the text instance. The orientation of the short line is the orientation of this R-anchor, while the length of the short line indicates the level of confidence. We can observe that the brighter field of the feature map focuses on the text region, while the other region becomes darker after 150,000 iterations. Moreover, the orientations of the regions become closer to the orientation of the text instance as the number of iterations increases.

\subsection{Accurate Proposal Refinement}

\begin{algorithm}[t]
    \caption{IoU computation}
    \begin{algorithmic}[1]
            \State {\textbf{Input}}: Rectangles $R_1,R_2,...,R_N$
            \State {\textbf{Output}}: IoU between rectangle pairs $IoU$
            \For {each pair $\langle R_i,R_j\rangle$ ($i<j$)}
            \State Point set $PSet\leftarrow \emptyset$
            \State Add intersection points of $R_i$ and $R_j$ to $PSet$
            \State Add the vertices of $R_i$ inside $R_j$ to $PSet$
            \State Add the vertices of $R_j$ inside $R_i$ to $PSet$
            \State Sort $PSet$ into anticlockwise order
            \State Compute intersection $I$ of $PSet$ by triangulation
            \State $IoU[i,j] \leftarrow$ $\frac{Area(I)}{Area(R_i) + Area(R_j) - Area(I)}$
            \EndFor
    \end{algorithmic}
    \label{alg:iou}
\end{algorithm}

\begin{figure}[t]
  \centering
  \includegraphics[width=\linewidth]{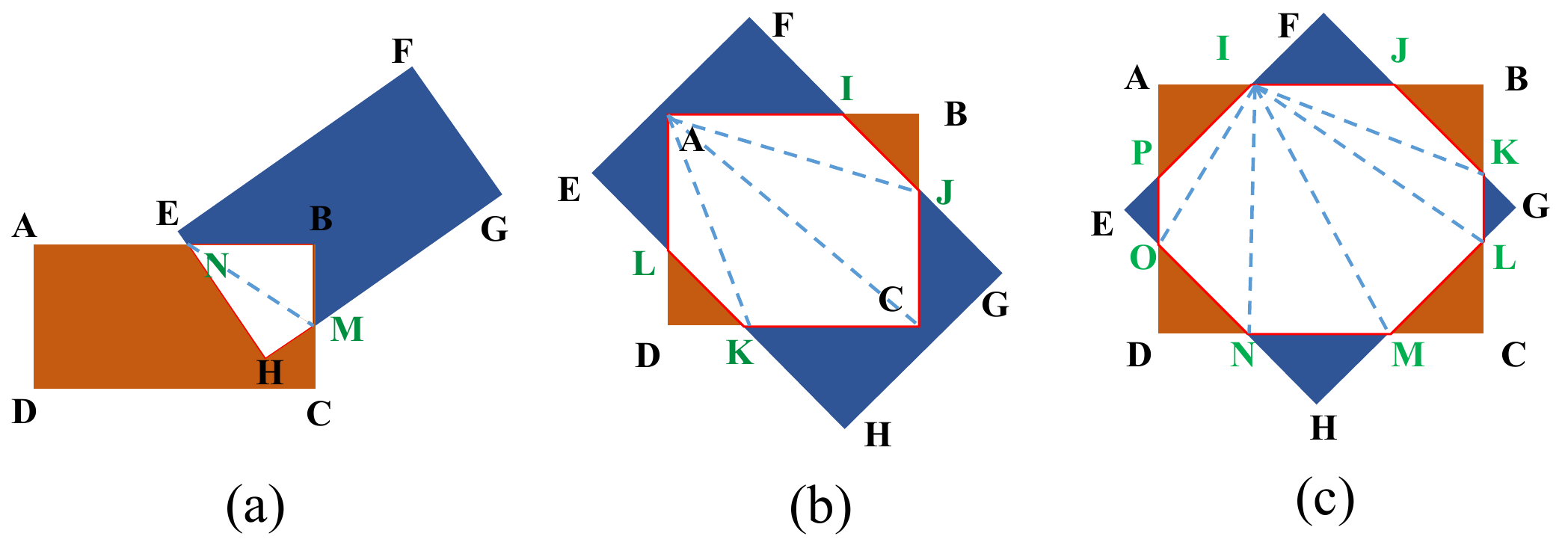}
  \caption{Examples of skew IoU computation: (a) 4 points, (b) 6 points, (c) 8 points (vertices of rectangle are in black, while intersection points are in green).
  Considering example (b), first add intersection points I, J, L, and K and inner vertices A and C to $PSet$, sort $PSet$ to obtain convex polygon AIJCKL, and then calculate the intersection area $Area$(AIJCKL) = $Area$($\Delta$AIJ) + $Area$($\Delta$AJC)+ $Area$($\Delta$ACK)+ $Area$($\Delta$AKL).}
  \label{fig:IoU}
\end{figure}

\textbf{Skew IoU Computation} The rotation proposals can be generated in any orientations. Thus, the IoU computation for axis-aligned proposals may lead to an inaccurate IoU of skew interactive proposals and further ruin the proposal learning. As shown in Algorithm \ref{alg:iou}, we design an implementation\footnote{Here, we use the GPU to accelerate the computation speed.} for the skew IoU computation with consideration of the triangulation \cite{Plaisted1987A}; Figure \ref{fig:IoU} shows the geometric principles.
Given a set of skew rectangles $R_1, ..., R_n$, our goal is to compute the IoU for each pair $\langle R_i,R_j\rangle$. The first step is to generate the intersection point set $PSet$ of $R_i$ and $R_j$ (Lines 4-7 in Algorithm 1). The intersection points of the two rectangles and the vertices of one rectangle inside another rectangle are calculated and inserted into $PSet$. Then, the intersection area of $PSet$ is computed (Lines 8-10 in Algorithm 1). The points in $PSet$ are sorted into anticlockwise order according to their positions in the image, and a convex polygon is generated based on the ordered points. By the triangulation, we can obtain the triangle set (e.g., \{$\Delta$AIJ, $\Delta$AJC, $\Delta$ACK, $\Delta$AKL\} in Figure \ref{fig:IoU}(b)). The area of the polygon is the sum of the areas of the triangles. Finally, the IoU value is computed.

\textbf{Skew Non-Maximum Suppression (Skew-NMS)} Traditional NMS takes only the IoU factor into consideration (\eg, the IoU threshold is $0.7$), but it is insufficient for arbitrary-oriented proposals. For instance, an anchor with a ratio of 1:8 and an angle difference of $\frac{\pi}{12}$ has an IoU of 0.31, which is less than 0.7; however, it may be regarded as a positive sample. Therefore, the Skew-NMS consists of 2 phases: (i) keep the max IoU for proposals with an IoU larger than $0.7$; (ii) if all proposals have an IoU in the range $[0.3, 0.7]$, keep the proposal with the minimum angle difference with respect to the ground truth (the angle difference should be less than $\frac{\pi}{12}$).

\subsection{RRoI Pooling Layer}

\begin{algorithm}[t]
    \caption{RRoI pooling}
    \begin{algorithmic}[1]
            \State {\textbf{Input}}: Proposal ($x,y,h,w,\theta$), pooled size ($H_r$,$W_r$), input feature map $InFeatMap$, spatial scale $SS$
            \State {\textbf{Output}}: Output feature map $OutFeatMap$
            \State $Grid_{w}, Grid_{h} \leftarrow \frac{w}{W_r}, \frac{h}{H_r}$
            \For {$\langle i, j\rangle$ $\in \{0,...,H_r-1\}\times\{0,...,W_r-1\}$ }
                \State $L, T \leftarrow x - \frac{w}{2} + jGrid_{w}, y - \frac{h}{2} + iGrid_{h}$
                \State $L_{rotate} \leftarrow (L-x)\cos\theta + (T-y)\sin\theta + x$
                \State $T_{rotate} \leftarrow (T-y)\cos\theta - (L-x)\sin\theta + y$
                \State $value \leftarrow 0$
                \For {$\langle k, l\rangle \in \{0,...,\lfloor Grid_h\cdot SS-1\rfloor\}\times\{0,...,\lfloor Grid_w\cdot SS-1\rfloor\}$}
                \State $P_x \leftarrow \lfloor L_{rotate}\cdot SS+l\cos\theta + k\sin\theta + \frac{1}{2}\rfloor$
                \State $P_y \leftarrow \lfloor T_{rotate}\cdot SS-l\sin\theta + k\cos\theta + \frac{1}{2}\rfloor$
                \If{$InFeatMap[P_y,P_x] > value$}
                    \State $value \leftarrow InFeatMap[P_y,P_x]$
                \EndIf
                \EndFor
                \State $OutFeatMap[i,j] \leftarrow value$
            \EndFor
    \end{algorithmic}
    \label{alg:rroi}
\end{algorithm}

\begin{figure}[t]
  \centering
  \includegraphics[width=.9\linewidth]{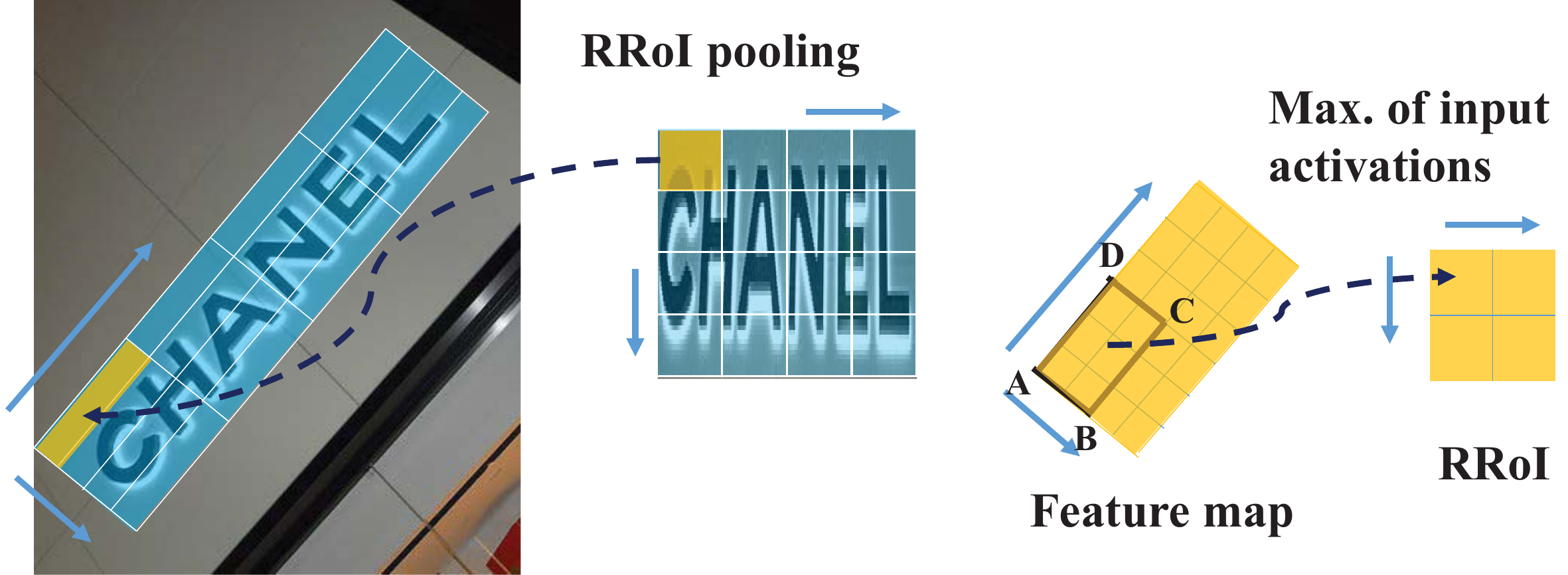}\\
  ~~~~~~(a)~~~~~~~~~~~~~~~~~~~~~~~~~~~~~~~~~(b)
  \caption{RRoI pooling layer: (a) divide arbitrary-oriented proposal into subregions; (b) max pooling of a single region from an inclined proposal to a point in the RRoI.}
  \label{fig:R-RoI}
\end{figure}

As presented for the Fast-RCNN \cite{girshick2015fast}, the RoI pooling layer extracts a fixed-length feature vector from the feature map for each proposal. Each feature vector is fed into fully connected layers that finally branch into the sibling \emph{cls} and \emph{reg} layers, and the outputs are the predicted localization and class of an object in an input image. As the feature map of image needs to be computed only once per image rather than computed for every generated proposal, the object detection framework is accelerated. The RoI pooling layer uses max pooling to convert the feature inside any valid RoI into a small feature map with a fixed spatial extent of $h_r \times w_r$, where $h_r$ and $w_r$ are layer hyperparameters that are independent of any RoI.

For the arbitrary-oriented text detection task, the traditional RoI pooling layer can only handle axis-aligned proposals. Thus, we present the rotation RoI (RRoI)
 pooling layer to adjust arbitrary-oriented proposals generated by RRPNs. We first set the RRoI layer hyperparameters to $H_r$ and $W_r$ for the RRoIs. The rotated proposal region can be divided into $H_r \times W_r$ subregions of $\frac{h}{H_r} \times \frac{w}{W_r}$ size for a proposal with height $h$ and width $w$ (as shown in Figure \ref{fig:R-RoI}(a)). Each subregion have the same orientation as that of the proposal. Figure \ref{fig:R-RoI}(b) displays an example with 4 vertices (A, B, C, and D) of the subregion on the feature map. The 4 vertices are calculated using a similarity transformation (shift, scale, and rotate) and grouped to range the border of the subregion. Then, max pooling is performed in every subregion, and max-pooled values are saved in the matrix of each RRoI; the pseudo-code for RRoI pooling is shown in Algorithm \ref{alg:rroi}. Compared with RoI pooling, RRoI pooling can pool any regions, with various angles, aspect ratios, or scales, into a fixed-size feature map. Finally, the proposals are transferred into RRoIs and sent to the classifiers to give the result, \ie, either text or background.

\section{Experiments}
\label{sec:evaluation}

We evaluate the rotation-based framework on three popular text detection benchmarks: MSRA-TD500 \cite{Yao2012Detecting}, ICDAR2015 \cite{karatzas2015icdar} and ICDAR2013 \cite{Karatzas2013ICDAR}. We follow the evaluation protocols of these benchmarks. The MSRA-TD500 dataset contains 300 training images and 200 testing images. Annotations of the images consist of both the position and orientation of each text instance, and the benchmark can be used to evaluate the text detection performance over the multi-oriented text instance. As the dataset of MSRA-TD500 is relatively smaller, its experiments are designed to exploit alternative settings. ICDAR2015 was released for the text localization of the incidental scene text challenge (Task 4.1) of the ICDAR 2015 Robust Reading Competition; it has 1,500 images in total.
Different from previous ICDAR robust reading competitions, the text instance annotations have four vertices, which form an irregular quadrilateral bounding box with orientation information. We roughly generate an inclined rectangle to fit the quadrangle and its orientation. The ICDAR2013 dataset is from the ICDAR 2013 Robust Reading Competition. There are 229 natural images for training and 233 natural images for testing. All the text instances in this dataset are horizontally aligned, and we conduct experiments on this horizontal benchmark to determine the adaptability of our approach to specific orientations.

\textbf{Implementation Details.} Our network is initialized by pretraining a model for ImageNet classification \cite{Simonyan2015Very}. The weights of the network are updated by using a learning rate of $10^{-3}$ for the first 200,000 iterations and $10^{-4}$ for the next 100,000 iterations, with a weight decay of $5\times 10^{-4}$ and a momentum of 0.9. We use the rotation of an image with a random angle for the data augmentation, as their efficiency and measurements are improved when the augmentation is used (see Table \ref{table:Rotation}).

\begin{table}[t]
\centering
\caption{Effect of data augmentation.}
\begin{tabular}{c|ccc}
\hline
Data Augmentation & Precision & Recall & F-measure \\\hline
  Without rotation &44.5\%&38.9\%&41.5\%\\\hline
  With rotation &{\bf 68.4\%}&{\bf 58.9\%}&{\bf 63.3\%}\\\hline
\end{tabular}
\label{table:Rotation}
\end{table}

Due to our different R-anchor strategy, the total number of proposals for each image is nearly 6 times that of previous approaches such as the Faster-RCNN. To ensure efficient detection, we filter the R-anchors to remove those passing through the border of an image. Therefore, the speed of our system is similar to that of previous works in both the training and testing stages; a comparison with the state-of-the-art approaches on MSRA-TD500 is presented in Table \ref{table:comparisonsall}-Left. Table \ref{table:Timing} shows the runtime speed of our proposed framework and that of the original Faster-RCNN under the baseline settings and with border padding. We can observe that our approach takes two times as much as the Faster-RCNN approach.

\begin{table}[t]
\centering
\caption{Runtime of proposed approach and of the Faster-RCNN. These runtimes were achieved using a single Nvidia Titan X GPU.}
\begin{tabular}{c|cc}
\hline
 & Baseline & With border padding \\\hline
  Faster-RCNN & 0.094 s & 0.112 s\\\hline
  The work & 0.214 s & 0.225 s\\\hline
\end{tabular}
\label{table:Timing}
\end{table}

\subsection{Ablation Study}
\label{sec:as}

\begin{figure}[t]
  \centering
  \includegraphics[width=\linewidth]{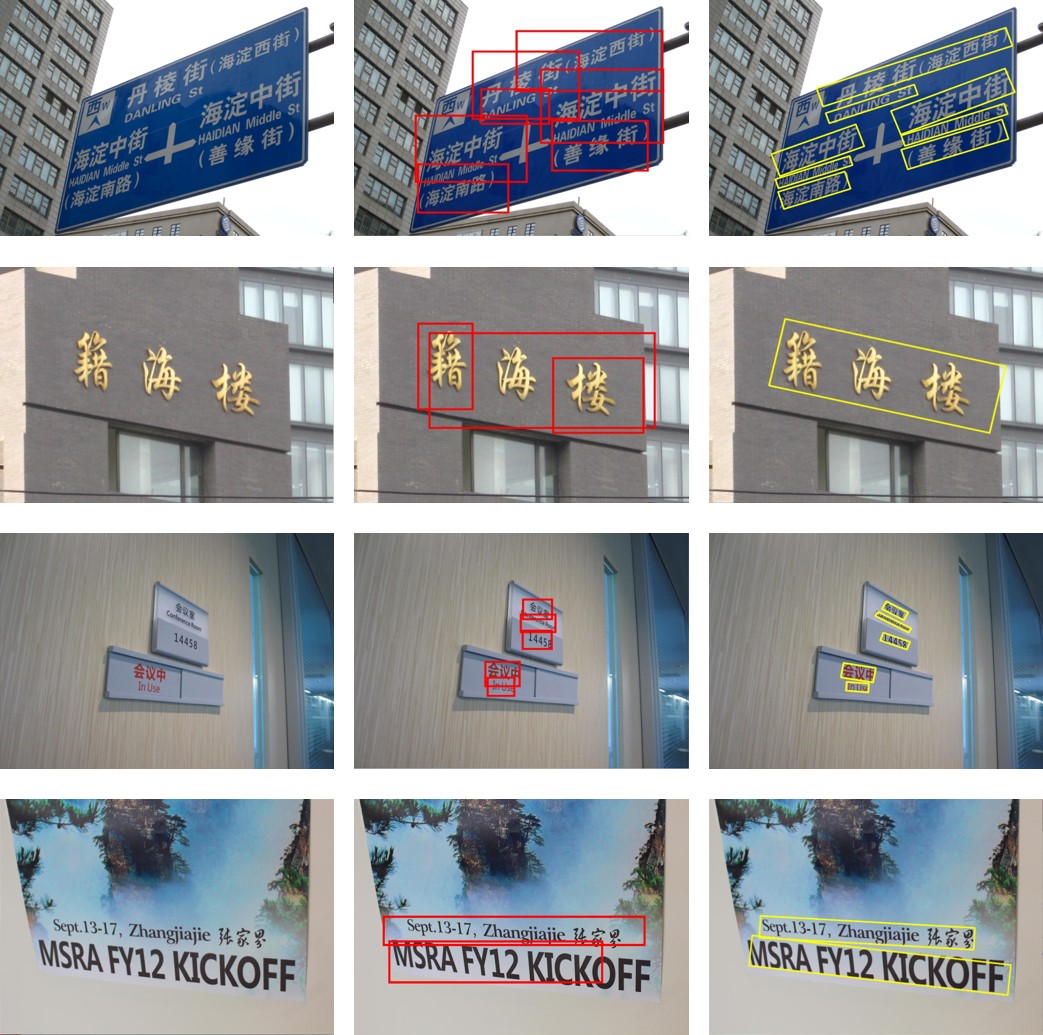}
  \caption{Comparison of rotation and horizontal region proposals. Left: original images; middle: text detection based on horizontal region proposal; right: text detection based on rotation region proposal.}
  \label{fig:rothor}
\end{figure}

\begin{table}[t]
\centering
\caption{Evaluation on MSRA-TD500 with different strategies and settings. Experiments on Faster-RCNN are based on the original source code. P, R and F denote the precision, recall, and F-measure, respectively. $\Delta$F is the improvement of the F-measure over the baseline. The strategies include the following: a. context of the text region; b. training dataset enlargement; c. border padding; and d. scale jittering.}
\begin{tabular}{|c|c|c|c|c|c|c|c|c|}
\hline
a. & b. & c. & d. & P & R & F & $\Delta$F\\\hline\hline
  \multicolumn{4}{|c|}{Faster-RCNN \cite{ren2016Faster}} & 38.7\%&30.4\%&34.0\%&-\\\hline
  \multicolumn{4}{|c|}{Baseline} & 57.4\%&54.5\%&55.9\%&-\\\hline
 $\surd$ & & & & 65.6\%&58.4\%&61.8\%&5.9\%\\\hline
 & $\surd$ & & & 63.3\%&58.5\%&60.8\%&4.9\%\\\hline
 & & $\surd$ & & 63.1\%&55.4\%&59.0\%&3.1\%\\\hline
 $\surd$ & $\surd$ & $\surd$ & & 68.4\%&58.9\%&63.3\%&7.4\%\\\hline
 $\surd$ & $\surd$ & $\surd$ & $\surd$ & {\bf 71.8\%}&{\bf 67.0\%}&{\bf 69.3\%}&13.4\%\\\hline
\end{tabular}
\label{table:baselineResult}
\end{table}

\begin{table}[t]
\centering
\caption{Exploitation of the text region context by enlarging the text bounding box by different factors of the original size.}
\begin{tabular}{c|ccc}
\hline
Factor & Precision & Recall & F-measure\\\hline
1.0&57.4\%&54.5\%&55.9\%\\
1.2&59.3\%&57.0\%&58.1\%\\
1.4&\textbf{65.6\%}&\textbf{58.4\%}&\textbf{61.8\%}\\
1.6&63.8\%&56.8\%&60.1\%\\\hline
\end{tabular}
\label{table:enlargement}
\end{table}

\begin{figure*}[t]
  \centering
  \includegraphics[width=.9\linewidth]{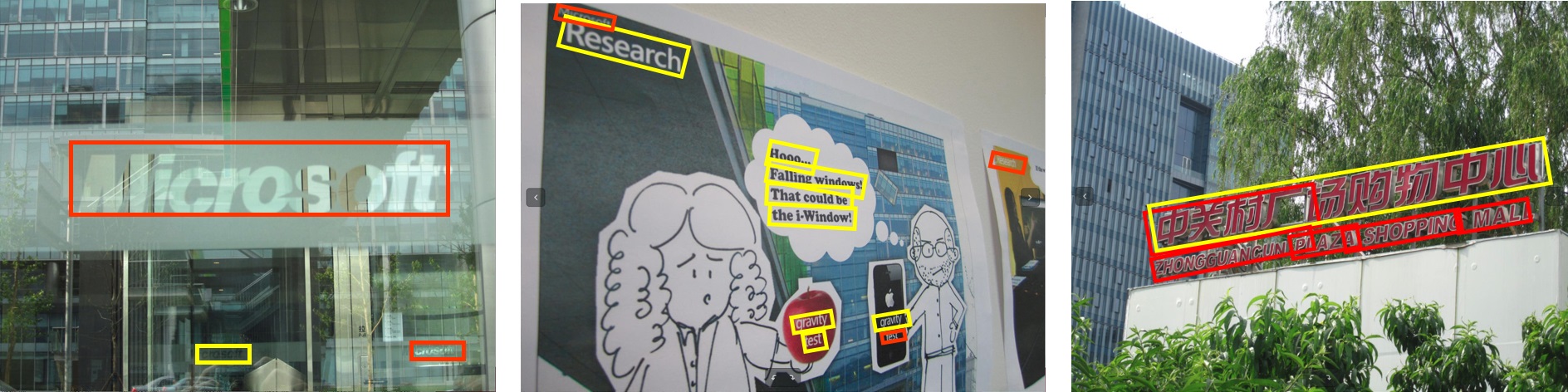}\\
  (a)~~~~~~~~~~~~~~~~~~~~~~~~~~~~~~~~~~~~~~~~~~~~(b)~~~~~~~~~~~~~~~~~~~~~~~~~~~~~~~~~~~~~~~~~~~~(c)\\
  \caption{Examples of failed detection on MSRA-TD500: (a) blur and uneven lighting situations; (b) extremely small text instances; (c) extremely long text line. The red boxes indicate instances of negative detection, \ie, either IoU$<$0.5 or failed to detect; the yellow boxes indicate instances of positive detection with respect to the ground truth.}
  \label{fig:failed}
\end{figure*}

We first perform an ablation study on the smaller dataset, \ie, MSRA-TD500. The baseline system is trained using 300 images from the MSRA-TD500 training set; the input image is resized, with long side being 1,000 pixels. The evaluation result is a precision of $57.4\%$, recall of $54.5\%$, and F-measure of $55.9\%$, which reflects a much better performance compared to that of the original Faster-RCNN, the P, R and F of which were $38.7\%$, $30.4\%$ and $34.0\%$, respectively. We make a comparison between rotation and horizontal region proposals, with some detection results illustrated in Figure \ref{fig:rothor}. The rotation-based approach is able to achieve accurate detection with less background area, which indicates the effectiveness of incorporating the rotation strategy.

Further analysis of the baseline results give us the following insights: (i) the difficult situations (\eg, blur and uneven lighting) in the image can hardly be detected; (ii) some text instances of extremely small size cannot be properly detected, resulting in a large recall loss regarding the performance; (iii) the extremely long text line, \ie, a height-width ratio of the bounding box larger than 1:10, cannot be correctly detected and is often split into several shorter proposals; hence, all the proposals become instances of false detection according to the evaluation of MSRA-TD500 (some failed detection instances are shown in Figure \ref{fig:failed}). A few alternative strategies and settings from the baseline approach are tested; a summary is given in Table \ref{table:baselineResult}.

\textbf{Context of the Text Region.} Incorporating the contextual information has been proven to be useful for the general object detection task (\eg, \cite{Redmon_2016_CVPR}), and we wonder whether it can promote a text detection system. We retain the center of the rotated bounding box and its orientation and enlarge both the width and height by a factor of $1.X$ in the data preprocessing step. During the testing phase, we divide the enlargement for every proposal. As shown in Table \ref{table:enlargement}, all the experiments exhibit an obvious increase in the F-measure. The reason may be that as the bounding box becomes larger, more context information of the text instance is obtained, and the information regarding the orientation can be better captured. Thus, the orientation of the proposals can be more precisely predicted.

\textbf{Training Dataset Enlargement.} We adopt HUST-TR400 (contains 400 images, with text instances annotated using the same parameters as for MSRA-TD500)~\cite{Yao2014A} as an additional dataset and form a training set of 700 images from both datasets. There is a significant improvement in all the measurements, and the F-measure is $60.8\%$, showing that the network is better trained and more robust when addressing noisy inputs.

\textbf{Border Padding.} Using our filtering strategy, most of the boundary breaking R-anchors are eliminated. However, as the bounding box is rotated by certain angles, it may still exceed the image border, especially when we enlarge the text region for the contextual information. Thus, we set a border padding of 0.25 times each side to reserve more positive proposals. The experiment shows that adding border padding to an image improves the detection results. The border padding increases the amount of computation for our approach by approximately 5\% (Table \ref{table:Timing}). In addition, combining border padding with enlargement of the text region and the training dataset yields a further improvement in the F-measure of $63.3\%$.

\begin{figure}[t]
  \centering
  \includegraphics[width=0.9\linewidth]{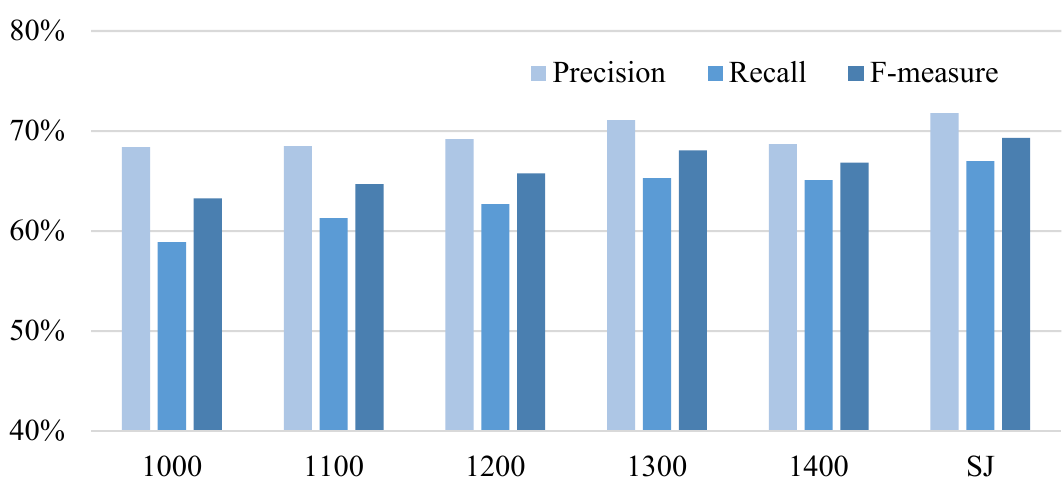}
  \caption{Evaluation on MSRA-TD500 for different input scales. $1X00$ ($X$=0,1,2,3,4) denotes those inputs with a long side of $1X00$ pixels, and SJ is the result with scale jittering. The experiments are conducted using strategies of context of the text region, training dataset enlargement, and border padding.}
  \label{fig:ScaleResult}
\end{figure}

\textbf{Scale Jittering.} There are still a number of small text regions in both training datasets, and we would like to improve the robustness of our system. One approach is to rescale the input images to a fixed larger size, and another is to perform scale jittering, \ie, rescaling with a long side of a random size before sending the image into the network. Figure \ref{fig:ScaleResult} shows that the inputs with a long side of 1300 pixels outperform those with other fixed settings (precision: 71.1\%, recall: 65.3\%, and F-measure: 68.1\%). When we apply scale jittering with a long side of a random size less than 1300 pixels, a better result is achieved compared to that of the experiment without jittering.

\begin{table}[t]
\centering
\caption{Training set and results for ICDAR2015. I\emph{XX} indicate ICDAR20\emph{XX} training set, M500 indicates MSRA-TD500, SVT indicates SVT dataset \cite{wang2010word}, and CA indicates data collected by the authors of \cite{tian2016detecting}.}
\begin{tabular}{|c||c|c|c|c|c|}
\hline
\small{Approach} &\multicolumn{2}{|c|}{RRPN} & \cite{yao2016scene} & \cite{tian2016detecting} & \cite{zhang2016multi}\\\hline\hline
\# of images & 2077 & 1229 & 1529& 3000& 1529\\\hline
Training & I13 I15 & I13 I15 & I13 I15 & I13 & I13 I15\\
set & I03 SVT &  & M500 & CA & M500\\\hline
\small{Precision} & {\bf 82.17\%} & 79.41\% & 72.26\%& 74.22\%& 70.81\%\\\hline
\small{Recall} & {\bf 73.23\%} & 70.00\% & 58.69\%& 51.56\%& 43.09\%\\\hline
\small{F-measure} & {\bf 77.44\%} & 74.41\% & 64.77\%& 60.85\%& 53.58\%\\\hline
\end{tabular}
\label{table:trainset}
\end{table}

\subsection{Performance on Benchmarks}
\textbf{MSRA-TD500.}
We use the best settings from the ablation study. The annotation of MSRA-TD500 prefers to label the region of a whole text line. Thus, the length of a text line does not have a fixed range, sometimes being very long. However, the ratios for the R-anchor are fixed and may not be large enough to cover all the lengths, which leads to several short bounding box results for a single text region. To address this extremely long text line issue, a post-processing step is incorporated by linking multiple short detection segments into a finer proposal, as detailed in Algorithm \ref{alg:post}. With this post-processing, the performance is further boosted, with the F-measure being 74.2\% and the time cost being only 0.3 s.
We also conduct an experiment that incorporates the post-processing as well as the strategies presented in Section \ref{sec:as} on the Faster-RCNN~\cite{ren2016Faster}; the results are a precision of 42.7\%, recall of 37.6\%, and F-measure of 40.0\%. The comparison verifies that using a rotation-based framework is necessary to achieve a more robust text detector.
Note that the post-processing is applied for the text line detection benchmark, \ie, MSRA-TD500, only and that we do not apply the algorithm to the ICDAR benchmarks. The results (RRPN) on MSRA-TD500 are shown in the left-most column of Table \ref{table:comparisonsall}.

\begin{algorithm}[t]
    \caption{Text-Linking Processing}
    \begin{algorithmic}[1]
        \State {\textbf{Input}}: Proposal $P_1,...,P_N~(P_k=x_k,y_k,h_k,w_k,\theta_k)$
        \State {\textbf{Output}}: Merged Proposal Set $PSet$
        \State Angle Threshold $T \leftarrow 10$
        \If{$ N ==1$}
            \State $PSet \leftarrow \{P_1\}$
        \EndIf
        \For {$k\in \{1,...,N\}$}
            \State Valid$[k] \leftarrow 1$
        \EndFor
        \For {each pair $\langle P_i,P_j\rangle$ ($i<j$)}
            \If{Valid$[i]==0$ or Valid$[j]==0$}
                \State Continue
            \EndIf
            \State MeanWidth $Width\leftarrow \frac{w_i + w_j}{2}$
            \State CenterDistance $Dis\leftarrow \sqrt{(x_i-x_j)^2 + (y_i-y_j)^2}$
            \State CenterGrad $ Grad \leftarrow |arctan(\frac{y_j-y_i}{x_j-x_i})|$
            \If{$ Dis < Width$ and $|Grad - \theta_i| < T$}
                \State $P_i \leftarrow \frac{x_i+x_j}{2},\frac{y_i+y_j}{2},\frac{h_i+h_j}{2},w_i+w_j,\frac{\theta_i+\theta_j}{2}$
                \State Valid$[j] \leftarrow 0$
            \EndIf
        \EndFor
        \State $PSet \leftarrow$ \{$P_k |$Valid$[k]$==1\}
    \end{algorithmic}
    \label{alg:post}
\end{algorithm}

\begin{figure}[t]
  \centering
  \includegraphics[width=.8\linewidth]{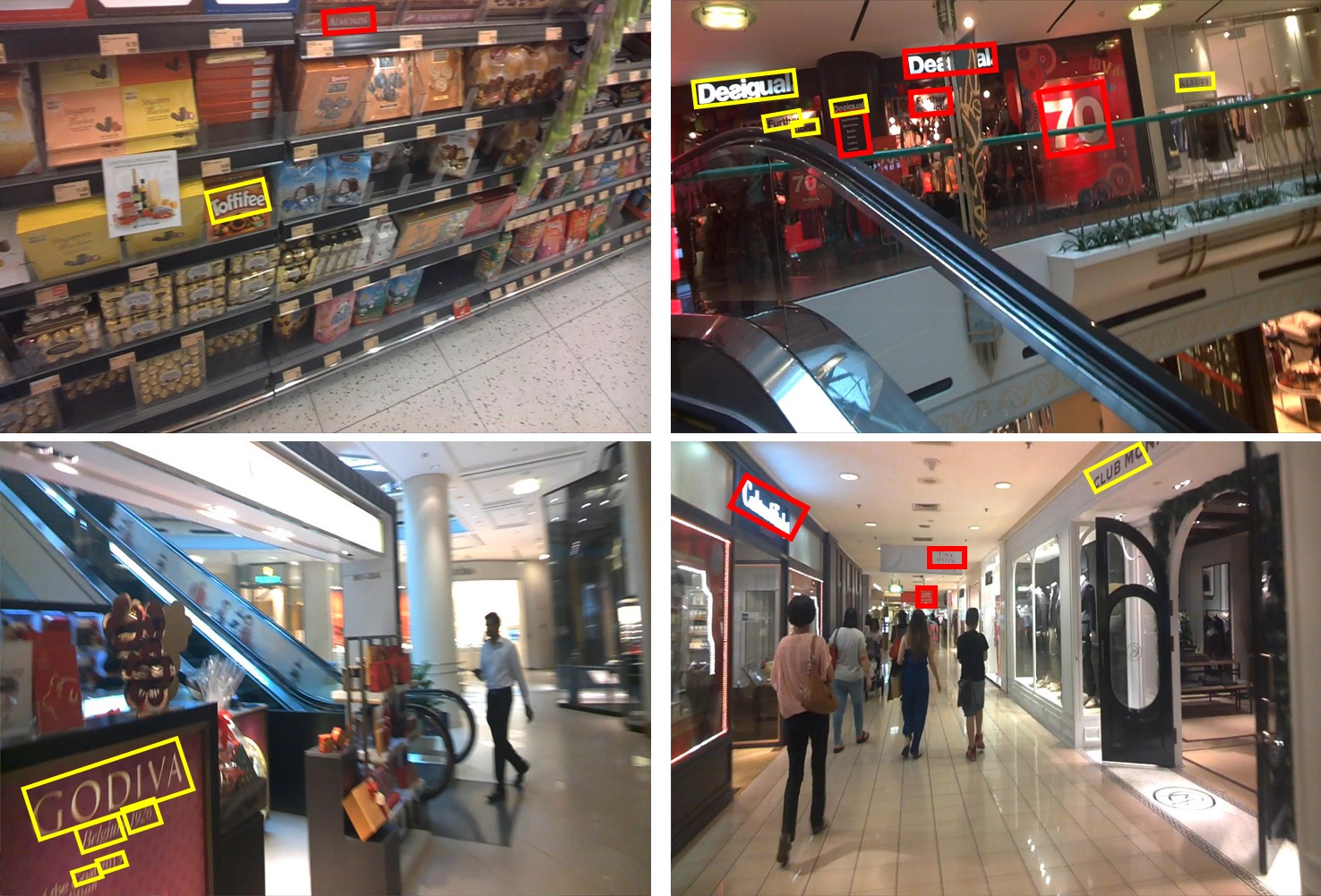}
  \caption{Text detection on ICDAR2015, with the model trained on the ICDAR2015 training set (including all unreadable text instances). The yellow areas denote instances of positive detection, with IoU $> 0.5$, while red areas represent text regions that were not detected.}
  \label{fig:falseDetection}
\end{figure}

\textbf{ICDAR 2015.} We train a baseline experiment on the ICDAR2015 benchmark using the same strategy used for MSRA-TD500. The evaluation result is a precision of 45.42\%, recall of 72.56\%, and F-measure of 55.87\%. There are some differences between these two datasets.
MSRA-TD500 tends to provide a text line ground truth, while ICDAR provides word-level annotations. Thus, the precision of our approach is lower than that of other methods that achieve the same F-measure. This issue may originate from three aspects.
First, some of the incidental text regions are still too small for our detector to find.
Second, there exist some small unreadable text instances (labeled `\#\#\#') in the ICDAR2015 training set, which may lead to the false detection of text-like instances.
Finally, our training set is insufficient (containing only 1,000 images) compared with those of previous approaches, such as \cite{yao2016scene,tian2016detecting,zhang2016multi}.  Some of the detection results obtained based on the ICDAR2015 training set are shown in Figure \ref{fig:falseDetection}.

\begin{figure}[t]
  \centering
  \includegraphics[width=.6\linewidth]{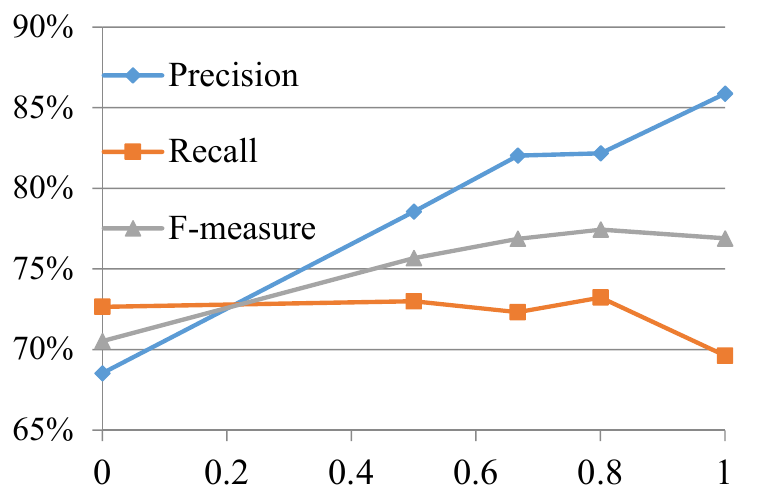}~~~
  \caption{Effect of unreadable text instance proportion using cross-validation on the training set. Horizontal axis represents the proportion of unreadable text instances removed; vertical axis represents the F-measure as a percentage.}
  \label{fig:proportion}
\end{figure}

To address the small text region issue, we create a larger scale by jittering the image patch with a long side of a random size less than 1,700 pixels before sending it into the network. We also check the impact of small unreadable text instances by randomly removing these instances from the training set. Figure \ref{fig:proportion} displays the curves of the measurements. The recall rate remains the same, \ie, approximately $72\%$-$73\%$, unless we remove all the unreadable instances, while the precision significantly increases with the proportion. Therefore, we randomly remove 80\% of the unreadable text instances in the training set and keep the whole testing set. To further improve our detection system, we incorporate a few text datasets for training, \ie, ICDAR2013 \cite{Karatzas2013ICDAR}, ICDAR2003 \cite{lucas2003icdar} and SVT \cite{wang2010word}. As listed in Table \ref{table:trainset}, the training images for different approaches are of the same order of magnitude, and ours achieves better performance.

\begin{table*}[t]
\centering
\caption{Comparison with state-of-the-art approaches on three benchmarks. Faster-RCNN results based on ICDAR2013 are reported in \cite{Zhong2016DeepText}. Bold text denotes the top result, while underlined text corresponds to the second runner-up.}
\begin{tabular}{c|cccc||c|ccc||c|ccc}
\hline
\multicolumn{5}{c||}{MSRA-TD500} & \multicolumn{4}{|c||}{ICDAR2015} & \multicolumn{4}{|c}{ICDAR2013}\\\hline
Approach & P & R & F & Time& Approach & P & R & F& Approach & P & R & F\\\hline
Yin \etal \cite{Yin2014Robust}&71&61&65&0.8 s &  CTPN \cite{tian2016detecting} &  74 & 52  & 61  &  Faster-RCNN \cite{ren2016Faster} &75&71&73  \\
Kang \etal \cite{Kang2014Orientation}&71&62&66 &- & Yao \etal \cite{yao2016scene}  & 72  &  59 & 65  &  Gupta \etal \cite{2016cvpr_Gupta}& 92 &76 & 83 \\
Yin \etal \cite{Yin2015Multi}&81&63&71 &1.4 s & SCUT\_DMPNet \cite{liu2017deep}  &68&73&71 &  Yao \etal \cite{yao2016scene} &89&80&84  \\
Zhang \etal \cite{zhang2016multi}&{\bf 83}&67&74& 2.1 s & UCSC\_TextSpotter \cite{qin2017cascaded} &65 &{\bf 79} &71 &  DeepText \cite{Zhong2016DeepText}&85&81&85   \\
Yao \etal \cite{yao2016scene}&77&{\bf 75}&{\bf 76}&0.6 s & hust\_orientedText \cite{shi2017detecting}  &  77 & 75  & 76  &   CTPN \cite{tian2016detecting} &\underline{93}&\underline{83}&\underline{88}  \\\hline
RRPN &82&68&74&0.3 s & RRPN & \underline{82} & 73 & \underline{77} & RRPN & 90 & 72 & 80 \\
RRPN* &\underline{82}&\underline{69}&\underline{75}&0.3 s & RRPN* & {\bf 84} & \underline{77} & {\bf 80} & RRPN* & {\bf 95} & {\bf 88} & {\bf 91} \\\hline
\end{tabular}

\label{table:comparisonsall}
\end{table*}

\begin{figure*}[t]
  \centering
  \includegraphics[width=\linewidth]{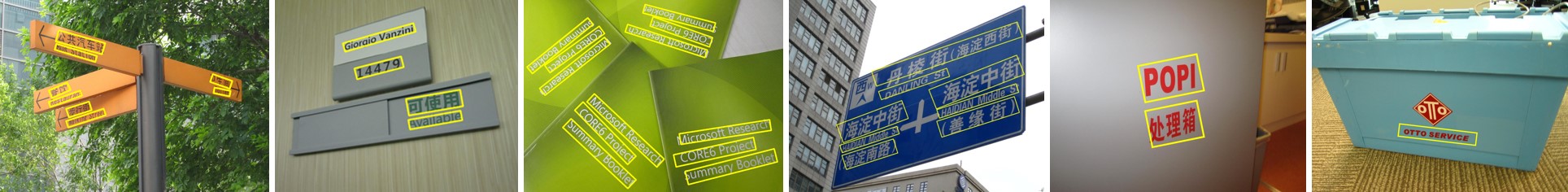}\\
  (a) MSRA-TD500\\
  \includegraphics[width=\linewidth]{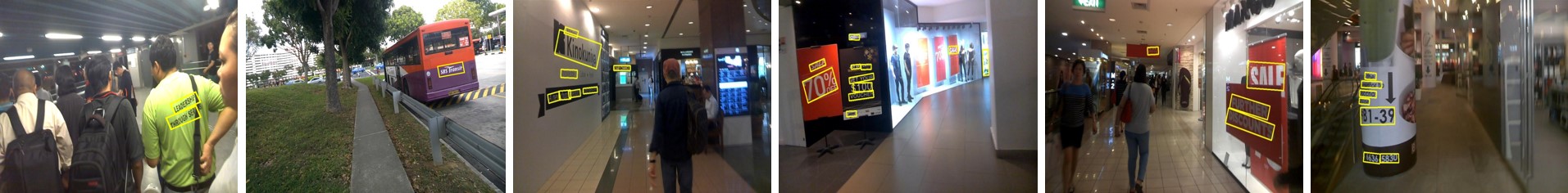}\\
  (b) ICDAR2015\\
  \includegraphics[width=\linewidth]{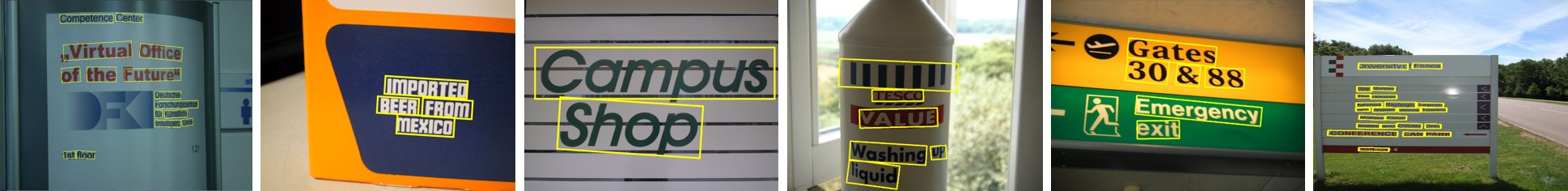}\\
  (c) ICDAR2013\\
  \caption{Text detection results for different benchmarks.}
  \label{fig:exampleall}
\end{figure*}

\textbf{ICDAR 2013.} To examine the adaptability of our approach, we also conduct experiments on the horizontal-based ICDAR2013 benchmark.
We reuse the model trained for ICDAR2015, and the 5-tuple rotation proposals are fit into horizontal-aligned rectangles.
The result is a precision of 90.22\%, recall of 71.89\%, and F-measure of 80.02\% under the ICDAR 2013 evaluation protocol. As shown in Table \ref{table:comparisonsall}, there is a 7\% improvement compared with the Faster-RCNN, which confirms the robustness of our detection framework with the rotation factor.

\subsection{More Results}

\begin{figure}[t]
  \centering
  \includegraphics[width=\linewidth]{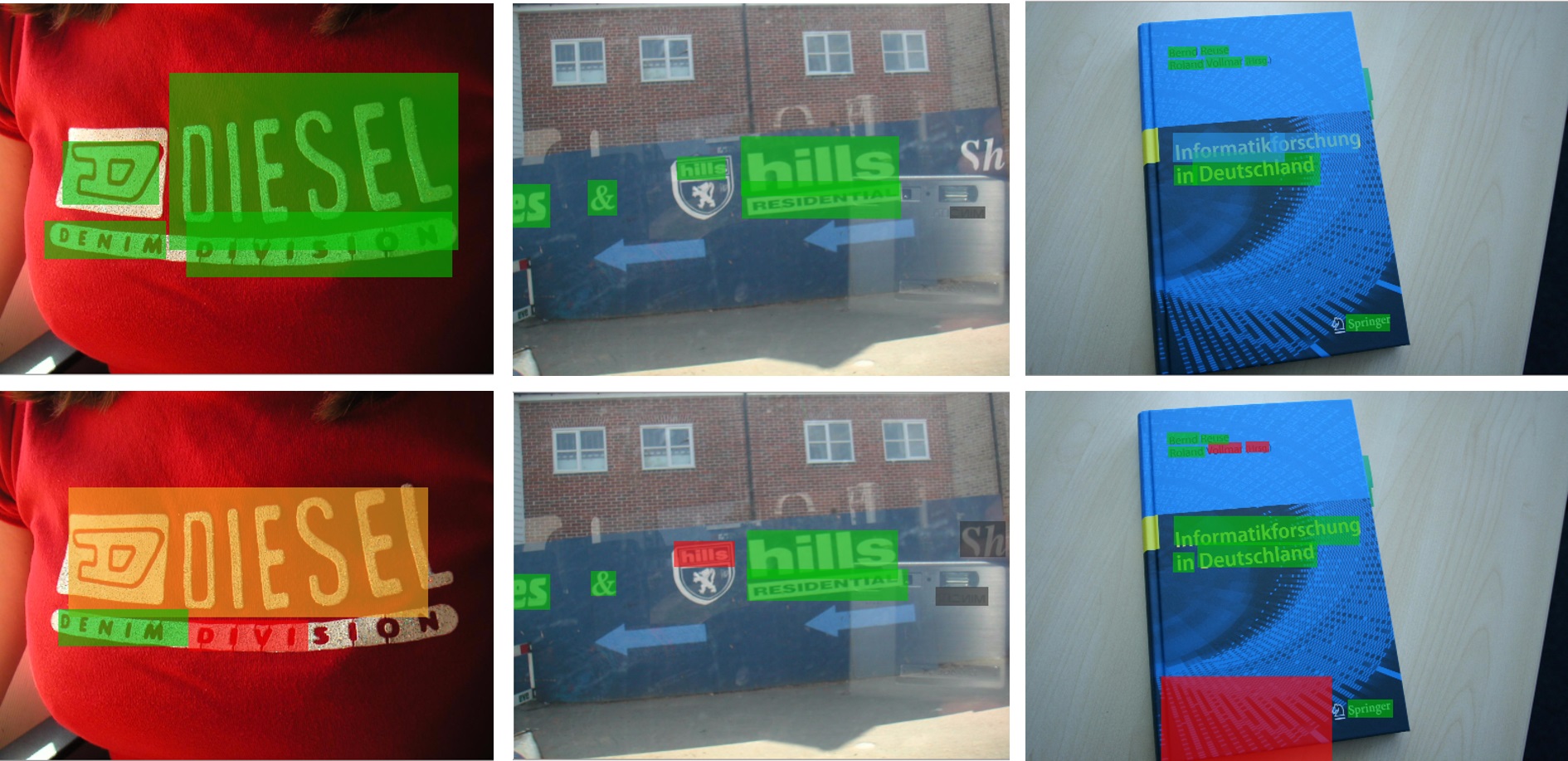}
  \caption{Detection results of the proposed approach and DeepText \cite{Zhong2016DeepText}, downloaded from the ICDAR evaluation website\protect\footnotemark[2]. The green and red boxes indicate instances of positive and false detection, respectively, the orange box refers to ``one box covering multiple instances'', and the blue box indicates multiple occurrences of detection for one instance.}
  \label{fig:deeptext}
\end{figure}
\footnotetext[2]{RRPN: \url{http://rrc.cvc.uab.es/?ch=2&com=evaluation&view=method_samples&task=1&m=15904&gtv=1}; DeepText: \url{http://rrc.cvc.uab.es/?ch=2&com=evaluation&view=method_samples&task=1&m=8665&gtv=1}}

\begin{figure}[t]
  \centering
  \includegraphics[width=\linewidth]{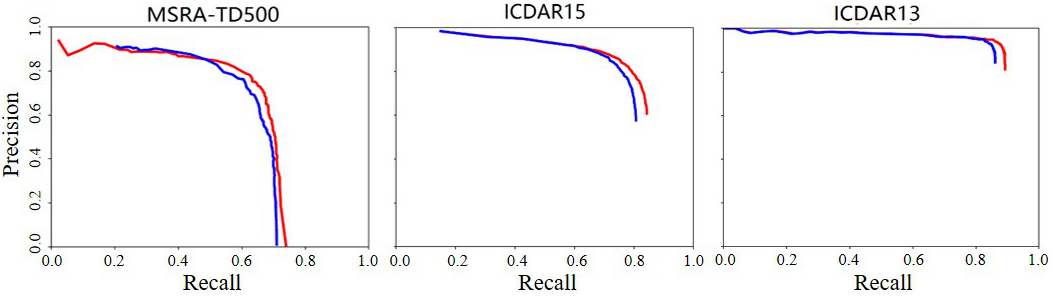}
  \caption{Precision-recall curve of the benchmarks. The red and blue curves represent the results of RRPN and RRPN*, respectively.}
  \label{fig:prcurve}
\end{figure}

The experimental results of our method compared with those of the state-of-the-art approaches are given in Table \ref{table:comparisonsall}.
As the RRPN models are trained separately for MSRA-TD500 and ICDAR, we also train a unified model (RRPN*) trained on all of the training sets to consider the generalization issue.
The precision-recall curves of RRPN and RRPN* on the three datasets are illustrated in Figure \ref{fig:prcurve}.
For the MSRA-TD500 dataset, the performance of our RRPN reaches the same magnitude of that of the state-of-the-art approaches, such as \cite{yao2016scene} and \cite{zhang2016multi}. When our system achieves text detection, it is more efficient than others, requiring a processing time of only 0.3 s per testing image.
For the ICDAR benchmarks, the substantial performance gains over the published works confirm the effectiveness of using a rotation region proposal and rotation RoI for the text detection task.
The recently developed DeepText \cite{Zhong2016DeepText} is also a detection-based approach, but it is based on the Inception-RPN structure. Both our approach and DeepText are evaluated on the ICDAR2013 benchmark. The evaluation results in Table \ref{table:comparisonsall} and detection examples in Figure \ref{fig:deeptext} demonstrate that our approach performs better in terms of different evaluation measurements. We believe that our rotation-based framework is also complementary to the Inception-RPN structure, as they both focus on different levels of information.
Some detection results obtained on the benchmarks are illustrated in Figure \ref{fig:exampleall}, and we have released the code and trained models for future research\footnotemark[3]\footnotetext[3]{\url{https://github.com/mjq11302010044/RRPN}}.

\section{Conclusions}
\label{sec:conclusion}

In this paper, we introduced a rotation-based detection framework for arbitrary-oriented text detection. Inclined rectangle proposals were generated with the text region orientation angle information from higher convolutional layers of network, resulting in the detection of text with multiple orientations. A novel RRoI pooling layer was also designed and adapted to the rotated RoIs. Experimental comparisons with the state-of-the-art approaches on MSRA-TD500, ICDAR2013 and ICDAR2015 showed the effectiveness and efficiency of our proposed RRPN and RRoI for the text detection task.

\ifCLASSOPTIONcaptionsoff
  \newpage
\fi

\bibliographystyle{IEEEtran}
\bibliography{total}

\end{document}